\documentclass[11pt]{article}

\usepackage[]{acl}

\usepackage{times, graphicx, footmisc, layouts}
\usepackage{latexsym,multirow}

\usepackage[utf8]{inputenc}

\usepackage{microtype}

\def\langCount{8}
\def\lrLangCount{7}
\def\goldenCount{5402} 
\def\totalDataCount{0.45M}
\title{XAlign: Cross-lingual Fact-to-Text Alignment and Generation for Low-Resource Languages}

\author{Tushar Abhishek, Shivprasad Sagare, Bhavyajeet Singh, Anubhav Sharma,\\ {\bf Manish Gupta\thanks{The
author is also a Principal Applied Scientist at Microsoft.}~ and Vasudeva Varma} \\ Information Retrieval and Extraction Lab, IIIT Hyderabad, India \\ \{tushar.abhishek,shivprasad.sagare,bhavyajeet.singh,anubhav.sharma\}@research.iiit.ac.in,\\ \{manish.gupta,vv\}@iiit.ac.in}

\begin{document}
\maketitle
\begin{abstract}
Multiple critical scenarios (like Wikipedia text generation given English Infoboxes) need automated generation of descriptive text in low resource (LR) languages from English fact triples. Previous work has focused on \emph{English} fact-to-text (F2T) generation. To the best of our knowledge, there has been no previous attempt on cross-lingual alignment or generation for LR languages. 
Building an effective \emph{cross-lingual F2T (XF2T)} system requires alignment between English structured facts and LR sentences.  
We propose two unsupervised methods for cross-lingual alignment. We contribute \textsc{XAlign}, an XF2T dataset with \totalDataCount{} pairs across \langCount{} languages, of which \goldenCount{} pairs have been manually annotated.
We also train strong baseline XF2T generation models on \textsc{XAlign} dataset.  
\end{abstract}

\section{Introduction}
Fact-to-text (F2T) generation~\cite{reiter1997building} is the task of transforming structured data into natural language. F2T systems are vital in many downstream applications like automated dialog systems, question answering, etc.
F2T generation requires a structured dataset which is well-aligned with semantically equivalent textual data. The manual creation of such a high-quality F2T dataset requires human supervision and is quite challenging to scale. Recently various automatic alignment approaches have been proposed like pairing up Wikipedia sentences with Infobox~\cite{lebret2016wikibio}, using distant supervision~\cite{elsahar2018trex}, finding the lexical overlap between textual and structural entities~\cite{jin2020genwiki, fu2020partially, agarwal2021knowledge}, etc.

\begin{figure}[h!]
    \centering
    \includegraphics[width=\columnwidth]{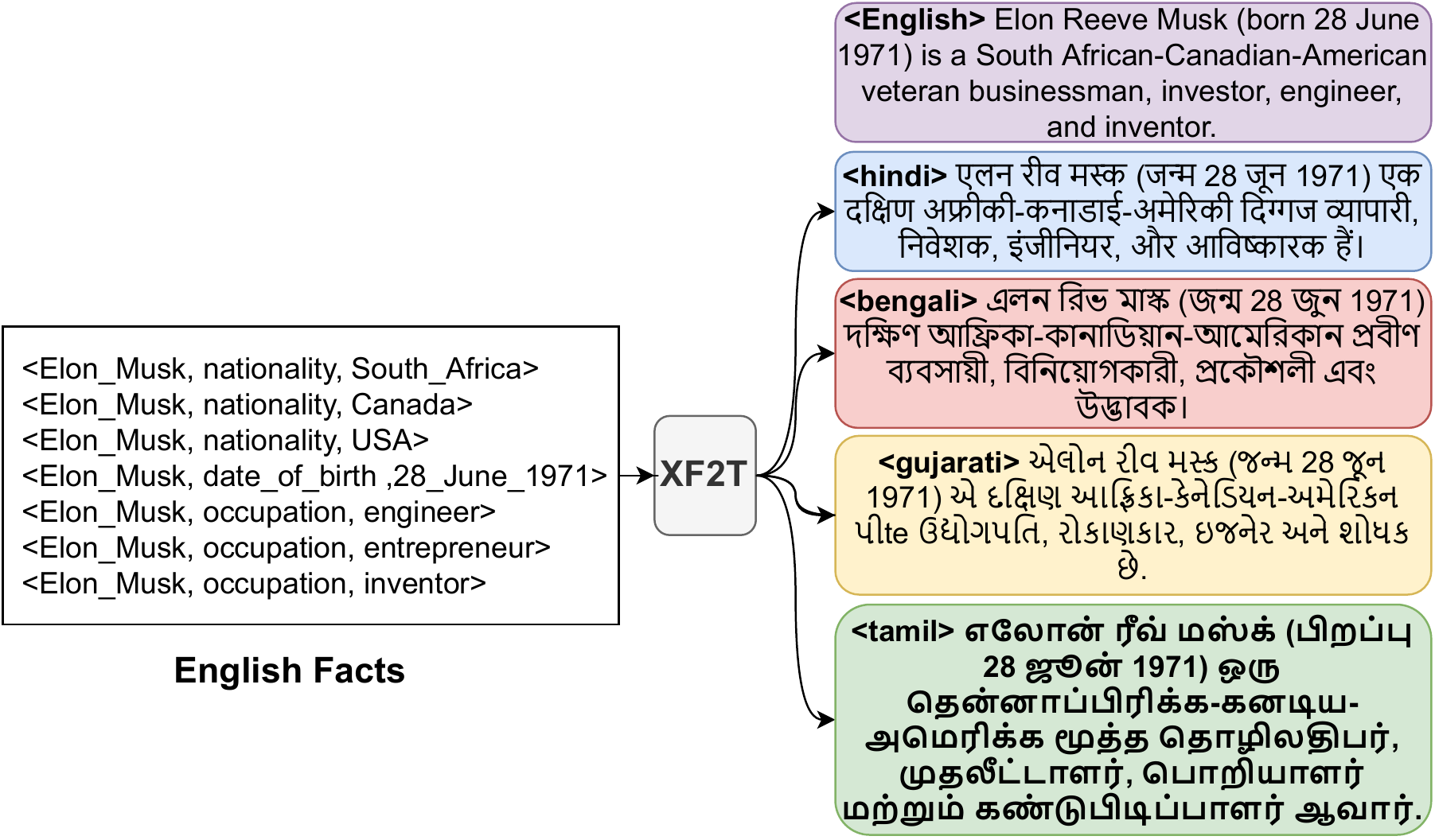}
    \caption{XF2T Example: Generating English, Hindi, Bengali, Gujarati or Tamil sentences to capture semantics from English facts.}
    \label{fig:examples}
\end{figure}

Most of the existing F2T datasets are English-only. For example, structured Wikidata entries for person entities in low resource (LR) languages are minuscule in number compared to that in English (Table~\ref{tab:wikiDataWikipediaStats} in Appendix). Also, average facts per entity in LR languages are much smaller than in English. Thus, monolingual F2T for LR languages suffers from data sparsity. In this work, we propose a novel task of XF2T generation which takes a set of English facts as input and generates a sentence capturing the fact-semantics in the specified language. For this task, we also contribute a new dataset, \textsc{XAlign}. Fig.~\ref{fig:examples} shows an XF2T example from our dataset. 




Overall, we make the following contributions. (1) We propose the problem of XF2T alignment and generation for LR languages. (2) We contribute a high quality XF2T dataset, \textsc{XAlign}, in \langCount{} languages with \goldenCount{} human-labeled examples. (3) We propose transfer learning and distant supervision based methods for cross-lingual alignment. (4) We train multiple multi-lingual XF2T models which lead to BLEU score of 25.02.

\begin{table*}[!t]
    \centering
    \scriptsize
    \begin{tabular}{|l|l|c|c|c|c|c|l|}
    \hline
Dataset&Languages&A/M&I&F&P&T&X-Lingual\\
\hline
\hline
WikiBio&en&A&728K&19.70&1740&26.1&No\\
\hline
E2E&en&M&50K&5.43&945&20.1&No\\
\hline
WebNLG 2017&en&M&25K&2.95&373&22.7&No\\
\hline
fr-de Bio&fr, de&A&170K, 50K&8.60, 12.6&1331, 1267&29.5, 26.4&No\\
\hline
TREX&en&A&6.4M&1.77&642&79.8&No\\
\hline
WebNLG 2020&en, ru&M&40K, 17K&2.68, 2.55&372, 226&23.7 &Yes\\
\hline
KELM&en&A&8M&2.02&663&21.2&No\\
\hline
WITA&en&A&55K&3.00&640&18.8&No\\
\hline
WikiTableT&en&A&1.5M&51.90&3K&115.9&No\\
\hline
GenWiki&en&A&1.3M&1.95&290&21.5&No\\
\hline
\hline
\textsc{XAlign}&en + \lrLangCount{} LR&A&0.45M&2.02&367&19.8&Yes\\
\hline
    \end{tabular}
    \caption{Statistics of popular F2T datasets: WikiBio~\cite{lebret2016wikibio}, E2E~\cite{novikova2017e2e}, WebNLG 2017~\cite{gardent2017webnlg}, WebNLG 2020~\cite{castro-ferreira-etal-2020-2020}, fr-de Bio~\cite{nema2018generating}, KELM~\cite{agarwal2021knowledge}, WITA~\cite{fu2020partially}, WikiTableT~\cite{chen2021wikitablet}, GenWiki~\cite{jin2020genwiki}, TREX~\cite{elsahar2018trex}, and XAlign (ours). Alignment method could be A (automatic) or M (manual). I=number of instances, F=avg fact count, P=number of unique predicates, T=avg word count.}
    \label{tab:dataStatsSurvey}
\end{table*}

\section{Related Work}
\label{sec:related}

We propose a new cross-lingual natural language generation (NLG) task: XF2T. From a knowledge graph (KG) and text linking perspective, as against entity linking (link mention in a sentence to a KG entity)~\cite{botha2020entity} and fact linking (linking sentence to a set of facts)~\cite{kolluru2021multilingual}, XF2T is the problem of generating a sentence given a set of facts. Table~\ref{tab:dataStatsSurvey} shows basic statistics of popular F2T datasets. Unlike other datasets which are mostly on English only, our dataset contains \langCount{} languages and is a cross-lingual dataset. 

Some previous studies~\cite{gardent2017webnlg} collected aligned data by crowdsourcing while others have performed automatic alignment~\cite{agarwal2021knowledge}. We explore two different unsupervised methods to perform cross-lingual alignment. Existing F2T methods can be classified as (1) template-based  methods~\cite{cimiano2013exploiting,duma2013generating}, (2) Seq-2-seq attention networks~\cite{lebret2016wikibio,mei2016talk,shahidi2020two}, (3) hierarchical attention networks~\cite{nema2018generating,chen2020kgpt}, (4) pretrained Transformer~\cite{vaswani2017attention} methods~\cite{ribeiro2021investigating,zhao2020bridging,chen2020kgpt,fu2020partially}. Template-based methods fail on fact triples in a previously unseen domain. Also, all of these methods focus on English F2T only. We focus on XF2T. We make our code and dataset publicly available\footnote{\url{https://github.com/tushar117/XAlign}\label{datafootnote}}, and hope that this will help advance further research in this critical area.

\section{Data Collection \& Pre-processing}
\label{sec:cross_creation}
The \textsc{XAlign} is an XF2T dataset which consists of sentences from LR language Wikipedia mapped to English fact triples from WikiData. It contains data for the following languages: Hindi (hi), Telugu (te), Bengali (bn), Gujarati (gu), Marathi (mr), Kannada (kn), Tamil (ta) and English (en).

We gather a list of 85K person entities from Wikidata which have a link to a corresponding Wikipedia page in at least one of our \lrLangCount{} LR languages. This leads to a dataset $D$ where every instance $d_i$ contains a tuple $\langle$entityID, English Wikidata facts, LR language, LR-language Wikipedia URL for the entityID$\rangle$.

We extract facts from the 20201221 WikiData dump for all the \langCount{} languages for each entity in $D$ using the WikiData API\footnote{\url{https://query.wikidata.org/}}. We gathered facts corresponding to only these Wikidata property (or relation) types which capture most of the useful factual information for person entities like WikibaseItem, Time, Quantity, Monolingualtext. We retain any additional supporting information associated with the fact triple as a fact qualifier. This leads to overall  $\sim$0.91M facts extracted for $\sim$85K entities across all the \langCount{} languages. 

For each language, we used the Wikiextractor~\cite{Wikiextractor2015} to extract text from the 20210520 Wikipedia xml dump. We split this main text into sentences using the Indic NLP Library~\cite{kunchukuttan2020indicnlp}, with a few additional heuristics to account for Indic punctuation characters, sentence delimiters and non-breaking prefixes. We prune out (1) other language sentences using Polyglot language detector\footnote{\url{https://polyglot.readthedocs.io/en/latest/Detection.html}}, (2)  sentences with less than 5 words or more than 100 words, (3) sentences which could potentially have no factual information (sentences with no noun or verb\footnote{For POS tagging, we used Stanza~\cite{qi2020stanza} for en, hi, mr, te, ta; LDC Bengali POS Tagger~\cite{bali2010indian} for bn; and ~\cite{patel2008part} for gu.}). For each sentence per Wikipedia URL, we also store the section information.

\section{F2T Alignment in \textsc{XAlign}}
\label{sec:alignment}
\begin{figure}[!t]
    \centering
    \includegraphics[width=\columnwidth]{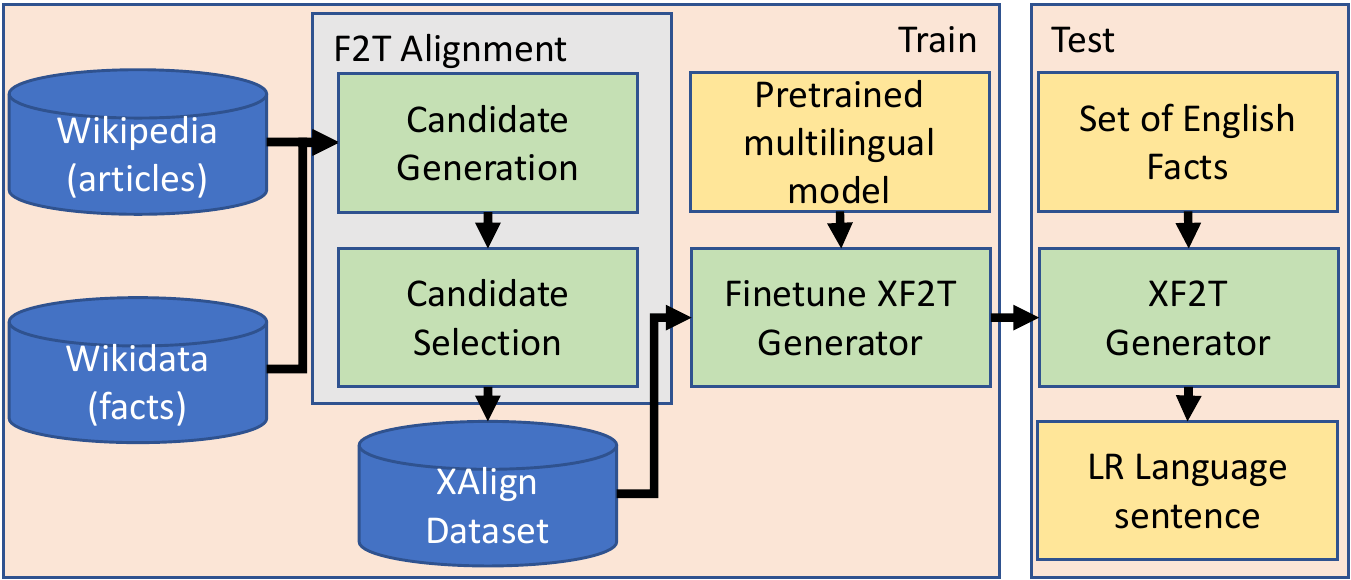}
    \caption{\textsc{XAlign}: XF2T System Architecture}
    \label{fig:system}
\end{figure}

For every (entity $e$, language $l$) pair, the dataset has a set $F_{el}$ of English Wikidata facts and a set of Wikipedia sentences $S_{el}$ in that language. Next, we build an automatic aligner to associate a sentence in $S_{el}$ with a subset of facts from $F_{el}$. As shown in Fig.~\ref{fig:system}, we propose a two-stage system. The first stage (Candidate Generation) generates (facts, sentence) candidates based on automated translation and syntactic+semantic match. The second stage (Selection) retains only those candidates which are strongly aligned using transfer learning and distant supervision.

\noindent\textbf{Stage 1: Candidate Generation}
Given a set of English facts $\{f_i\}_{i=1}^{|f|}$ and set of sentences $\{s_j\}_{j=1}^{|s|}$ in language $l$, we compute a similarity score that captures syntactic as well as semantic similarity between a (fact $f_i$, sentence $s_j$) pair.
For syntactic match, we use TFIDF by translating either the fact to language $l$ or the sentence to English. For semantic match, we compute cosine similarity between MuRIL~\cite{khanuja2021muril} representations of the fact and the sentence, or between their translations. We retain a sentence if the most similar fact has similarity score $>\tau$. For every sentence, we retain at most top-$K$ facts sorted according to their scores. We set $\tau=0.65$ and $K$=10 for our expts empirically.

\noindent\textbf{Manual Annotations for Ground-Truth Data}
We perform manual annotation in two phases. For both the phases, the annotators were presented with (LR language sentence $s$, $K$ English facts) output by Stage 1. They were asked to mark facts present in $s$. There were also specific guidelines to ignore redundant facts, handle abbreviations, etc. More detailed annotation guidelines and ethical statement are mentioned in the Appendix.

In the first phase, we got 60 instances labeled per language by a set of 8 expert annotators (trusted graduate students who understood the task very well). In phase 2, we first tested annotators using phase 1 data as golden control set. We selected 8 annotators per language from the National Register of Translators\footnote{\url{https://www.ntm.org.in/languages/english/nrtdb.aspx}}. Next, we choose up to 4 annotators per language who scored highest (on Kappa score with golden annotations) for final annotations. We report details of this dataset in Table~\ref{tab:annotationStats}. 

\begin{table}[!h]
    \centering
    \scriptsize
    \begin{tabular}{|l|c|c|c|c|c|}
    \hline
    Lang & $\kappa$&A&I&avg/min/max T&avg/min/max F\\
    \hline
    \hline
    hi&0.81&4&842&11.1/5/24&2.1/1/5\\
    \hline
    mr&0.61&4&736&12.7/6/40&2.1/1/8\\
    \hline
    te&0.56&2&734&9.7/5/30&2.2/1/6\\
    \hline
    ta&0.76&2&656&9.5/5/24&1.9/1/8\\
    \hline
    en&0.74&4&470&17.5/8/61&2.7/1/7\\
    \hline
    gu&0.50&3&530&12.7/6/31&2.1/1/6\\
    \hline
    bn&0.64&2&792&8.7/5/24&1.6/1/5\\
    \hline
    kn&0.54&4&642&10.4/6/45&2.2/1/7\\
    \hline
    \end{tabular}
    \caption{Annotation statistics of Ground Truth labeled test part of XAlign dataset. A=\#Annotators, I=\#instances, T=word count, F=fact count, $\kappa$=avg Kappa score}
    \label{tab:annotationStats}
\end{table}

\begin{table*}
    \centering
    \scriptsize
    \begin{tabular}{|p{0.6in}|l|c|c|c|c|c|c|c|c||c|}
    \hline
    Methods&&hi&mr&te&ta&en&gu&bn&kn&Avg.\\
    \hline
    \hline
    \multirow{3}{0.6in}{Baselines}&KELM-style~\cite{agarwal2021knowledge}&0.493&0.426&0.368&0.451&0.41&0.372&0.436&0.338&0.411\\
    \cline{2-11}
    &WITA-style~\cite{fu2020partially}&0.507&0.574&0.517&0.459&0.602&0.500&0.535&0.530&0.528\\
    \cline{2-11}
    &Stage-1 + TF-IDF&0.750&0.685&0.693&0.718&0.737&0.701&0.787&0.647&\textbf{0.715}\\
    \hline
    \hline
    \multirow{3}{0.6in}{Distant supervision}&MuRIL~\cite{khanuja2021muril}&0.763&0.684&0.74&0.755&0.705&0.785&0.624&0.677&0.717 \\
    \cline{2-11}
    &XLM-Roberta~\cite{conneau2020unsupervised}&0.781&0.69&0.765&0.739&0.765&0.785&0.669&0.724&0.740 \\
    \cline{2-11}
    &mT5~\cite{xue2021mt5}&0.790&0.714&0.776&0.786&0.766&0.800&0.698&0.705&\textbf{0.754}\\
    \hline
    \hline
    \multirow{3}{0.6in}{Transfer learning}&MuRIL~\cite{khanuja2021muril}&0.716&0.717&0.765&0.751&0.734&0.787&0.795&0.718&0.748 \\
    \cline{2-11}
    &XLM-Roberta~\cite{conneau2020unsupervised}&0.772&0.767&0.78&0.812&0.79&0.805&0.831&0.727&0.786 \\
    \cline{2-11}
    &mT5~\cite{xue2021mt5}&0.902&0.831&0.841&0.886&0.845&0.851&0.751&0.785&\textbf{0.837} \\
    \hline
    \end{tabular}
    \caption{Stage-2 (Fact, Sentence) Candidate Selection F1 Scores. For TF-IDF aligner, we used candidates generated from Stage-1. For KELM and WITA-style aligners, we followed method from their paper.}
    \label{tab:stage2Results}
\end{table*}

\noindent\textbf{Stage 2: Candidate Selection}
For every entity and language pair, Stage 1 outputs sentences each associated with a maximum of $K$ facts. We use two different techniques to retain only strongly aligned (fact, sentence) pairs: transfer learning from NLI (Natural language Inference) task and distant supervision from another English-only F2T dataset. For both methods, input=``sentence$\langle$SEP$\rangle$subject|predicate|object''.

\noindent Transfer learning from NLI: Given a premise sentence and a hypothesis sentence, NLI aims to predict whether the hypothesis entails, contradicts or is neutral to the given premise. Fact to sentence alignment is semantically similar to NLI where the sentence and the fact can be considered as the premise and the hypothesis respectively. We experiment with these multi-lingual NLI models: XLM-R, mT5, MuRIL. We use their Xtreme-XNLI finetuned checkpoints from Huggingface\footnote{MuRIL does not support vocabulary for all XNLI languages, so we finetuned it only for en, hi and ur.}. At inference time, if the model returns entailment as the prediction, we consider the (fact, sentence) pair to be aligned, else not. Thus, from amongst $K$ candidate facts for every sentence, we select a subset of facts, which are compared with the golden fact list for the given sentence to choose the best model.

\noindent Distant supervision: Given a (English fact, LR language sentence) pair, we train a binary classifier to predict whether the fact is associated with the LR language sentence or not, using the Knowledge Enhanced Language Modeling (KELM)~\cite{agarwal2021knowledge} dataset. We discuss details of processing the KELM dataset in the Appendix. Since our dataset is cross-lingual, for inference on output of the Stage 1 data, we experiment with cross-lingual, translate-test and translate-train settings. Translate-train  performs the best and hence we report results using this setting.

Table~\ref{tab:stage2Results} shows candidate selection F1 scores across all the languages  on our golden annotated dataset. We observe that the mT5 with transfer learning performs the best. 

\section{\textsc{XAlign} Dataset Analysis and XF2T Generation}
\label{sec:approach}
We run mT5 Stage-2 aligner on Stage-1 output to get the Train+Validation part of our \textsc{XAlign} Dataset.  Table~\ref{tab:xalignDataStats} shows basic dataset statistics. Figs.~\ref{fig:language_fact_number_fraction} and~\ref{fig:train_test_val_factcount} in the Appendix show fact count distribution.  Table~\ref{tab:topkpredicates} in the Appendix shows top 10 frequent fact properties across all the languages.

\begin{table}[!h]
    \centering
    \scriptsize
    \begin{tabular}{|l|c|c|c|c|c|}
    \hline
Lang &I&avg/min/max T&avg/min/max F&V\\
    \hline
    \hline
    hi&56582&25.3/5/99&2.0/1/10&75037\\
    \hline
    mr&19408&20.4/5/94&2.2/1/10&49512\\
    \hline
    te&24344&15.6/5/97&1.7/1/10&60865\\
    \hline
    ta&56707&16.7/5/97&1.8/1/10&121212\\
    \hline
    en&132584&20.2/4/86&2.2/1/10&103626\\
    \hline
    gu&9031&23.4/5/99&1.8/1/10&34605\\
    \hline
    bn&121216&19.3/5/99&2.0/1/10&130684\\
    \hline
    kn&25441&19.3/5/99&1.9/1/10&87760\\
    \hline
    \end{tabular}
    \caption{Basic Statistics of \textsc{XAlign} Train + Validation. I=number of instances, T=word count, F=fact count, V=Vocabulary size}
    \label{tab:xalignDataStats}
\end{table}

For XF2T generation, we train multiple  multi-lingual text generation models on Train+Validation part of our \textsc{XAlign} dataset. We use mT5-small, a basic Transformer model, GAT+Transformer model for the XF2T task. We also compare with a baseline where English facts are translated to LR language and concatenated to generate output. While translating if mapped strings for entities were present in Wikidata they were directly used. For each of these models, we also concatenate the facts with the section header information in the input text. We list hyper-parameter settings in Appendix. Table~\ref{tab:finalResults} shows BLEU results across different (model, language) combinations. Na\"ive translation baseline performs poorly. mT5 performs best with 25.0 avg BLEU across all languages. While the accuracy is good for en, hi and bn, further work needs to be done to improve XF2T for other LR languages. Table~\ref{tab:caseStudies} in Appendix shows XF2T prediction examples for our mT5 model.

\setlength{\tabcolsep}{3pt}
\begin{table}
    \centering
    \scriptsize
    \begin{tabular}{|l|c|c|c|c|c|c|c|c||c|}
\hline
Model&hi&mr&te&ta&en&gu&bn&kn&Avg.\\
\hline
\hline
Baseline&2.7&2.0&1.0&1.7&1.01&0.6&2.7&0.4&1.5\\
\hline
Transformer&35.4&17.3&6.9&8.8&38.8&13.2&35.6&3.1&19.9\\
\hline
GAT+Trans.&29.5&17.9&4.9&7.2&40.3&11.3&30.2&5.1&18.3\\
\hline
mT5&\textbf{40.6}&\textbf{20.2}&\textbf{11.4}&\textbf{13.6}&\textbf{43.7}&\textbf{16.6}&\textbf{45.3}&\textbf{8.7}&\textbf{25.0}\\
\hline
    \end{tabular}
    \caption{XF2T BLEU scores on \textsc{XAlign} test set}
    \label{tab:finalResults}
\end{table}

\section{Conclusion}
\label{sec:conclusion}
We propose a novel XF2T problem, contribute a new \textsc{XAlign} dataset for \langCount{} languages, propose two effective F2T alignment methods, and report BLEU results on strong baseline multi-lingual models. We strongly believe that our alignment models can significantly reduce human annotation costs in low-resource (LR) NLG. We also plan to extend this work to other LR languages.

\section*{Acknowledgements}
This research was partially funded by Ministry of Electronics and Information Technology (MeitY), Government of India under Sanction Order No: 11(6)/2019-HCC (TDIL). The views and conclusions contained herein are those of the authors and should not be interpreted as necessarily representing the official policies, either expressed or implied of MeitY.  

We would like to thank the following annotators from National Translation Mission for their crucial contributions in creation of test dataset: Bhaswati Bhattacharya, Aditi Sarkar, Raghunandan B. S., Satish M., Rashmi G.Rao, Vidyarashmi P N, Neelima Bhide, Anand Bapat, Rajendra Sagare, Krishna Rao N V, Nagalakshmi DV, Aditya Bhardwaj Vuppula, Nirupama Patel, Asir. T, Sneha Gupta, Dinesh Kumar and Jasmin Gilani.

\bibliography{main}
\bibliographystyle{acl_natbib}

\appendix
\section{Ethical Statement}
We do not foresee any harmful uses of this technology. In fact, F2T generation systems are vital in many downstream Natural Language Processing (NLP) applications like automated dialog systems~\cite{wen2016multi}, domain-specific chatbots~\cite{novikova2017e2e}, open domain question answering~\cite{chen2020kgpt}, authoring sports reports~\cite{chen2008learning}, etc. We believe that these systems will be useful for powering business applciations like Wikipedia text generation given English Infoboxes, automated generation of non-English product descriptions using English product attributes, etc.

As part of this work, we collected labeled data as discussed in Section~\ref{sec:alignment}. The dataset does not involve collection or storage of any personal identifiable information or offensive information at any stage. Human annotators were paid appropriately while performing data collection according to the standard wages set by National Translation Mission (\url{https://www.ntm.org.in/}) and mutually agreed upon.

\section{Wikidata+Wikipedia statistics across languages}
As shown in Table~\ref{tab:wikiDataWikipediaStats}, structured Wikidata entries for person entities in low resource (LR) languages are minuscule in number compared to that in English. Also, average facts per entity in LR languages are much smaller than in English.
\begin{table}[!b]
    \centering
    \scriptsize
    \begin{tabular}{|p{0.3in}|p{0.5in}|p{0.3in}|p{0.7in}|p{0.5in}|}
\hline
Lang.&WikiData entries&Facts&Average facts per entity&Wikipedia articles\\
\hline
\hline
hi&26.0K&271.0K&10.43&22.9K\\
\hline
mr&16.5K&174.0K&10.56&15.9K\\
\hline
te&12.4K&142.2K&11.49&7.8K\\
\hline
ta&26.0K&280.4K&10.77&25.6K\\
\hline
en&1.3M&30.2M&22.8&627.9K\\
\hline
gu&3.5K&37.8K&10.88&1.9K\\
\hline
bn&36.2K&501.9K&13.87&29.0K\\
\hline
kn&7.5K&83.6K&11.1&4.5K\\
\hline
    \end{tabular}
    \caption{WikiData+Wikipedia statistics for the person entities across languages}
    \label{tab:wikiDataWikipediaStats}
\end{table}

\section{Annotation Guidelines}
\subsection{Main Task}
The task is to mark English facts that are present in the given LR language sentence. You should choose all the facts that can be inferred from the given sentence by selecting the checkbox against it. Also, mention if the set of selected facts partially/completely cover the semantic information mentioned in the sentence.
\subsection{Instructions related to platform}
\begin{itemize}
    \item When you select a question, you will see a sentence in low resource (LR) language and a list of English facts.
\item Please read the LR sentence carefully. Although English translated sentence is provided for the reference, don't rely entirely on it. The translated sentence may not be accurate all the time.
\item You will find list of English facts below the sentence. Please choose the facts that can be inferred from the given sentence by selecting the checkbox against it.
\item If the sentence is grammatically incorrect,  incomplete or erroneous for any other reason, please mention the reason in the textbox at the bottom.
\end{itemize}

\begin{figure*}[!t]
    \centering
    \includegraphics[width=1.9\columnwidth]{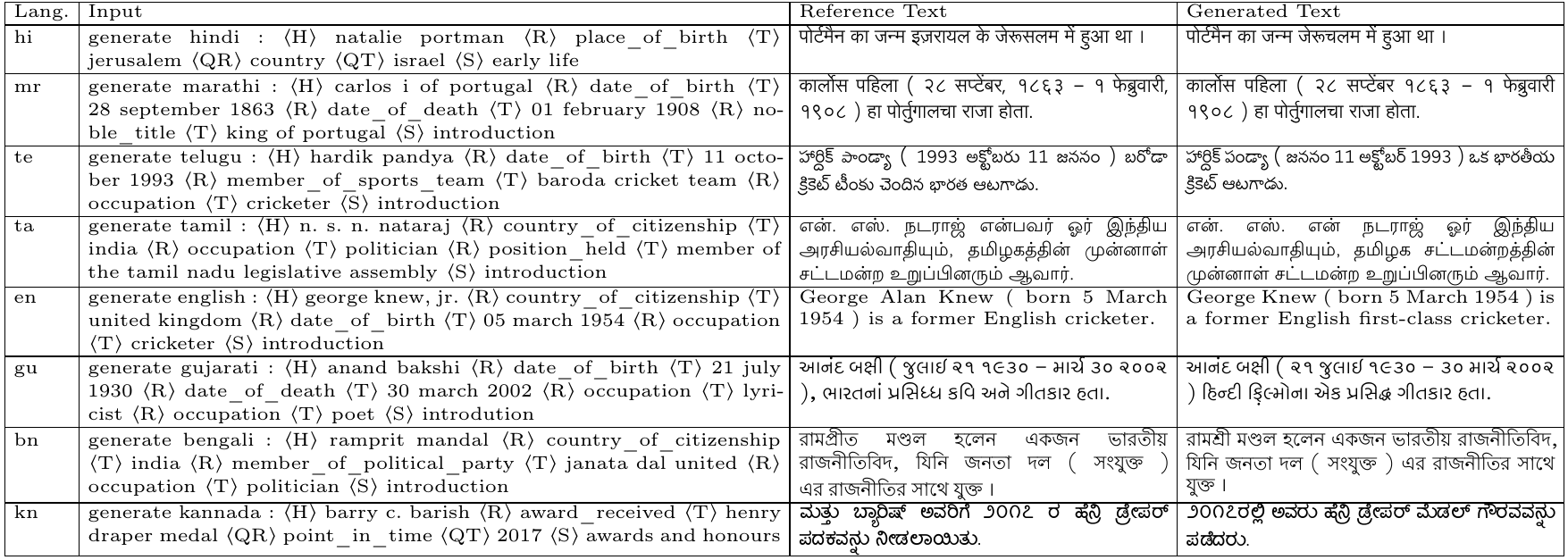}
    \captionof{table}{Test Dataset Examples with reference text and predictions from our mT5-small model.}
    \label{tab:caseStudies}
\end{figure*}

\subsection{Instructions related to annotations}
\begin{itemize}
\item  Exact Fact Matching: Information should exactly match what is present in the sentences (some exceptions are mentioned later; other than them, follow this rule strictly). For example, 
\begin{itemize}
\item Sentence: Tina Munim (DOB: 11 Feb 1955) is an actress who acts in Hindi movies.
\item Fact: Date of Birth | 11 February 1957. 
\item Although the fact mentions that date of birth is 11 Feb 1957 but we won't consider it as a valid alignment for the sentence.
\end{itemize}
\item Implied Information in facts
\begin{itemize}
\item If information is related to language related inference and does not require external world knowledge (a piece of knowledge not embedded in language itself), we mark that fact.
\begin{itemize}
\item Sentence: P. Nagarajan is a Member of Parliament in India's 16th Lok Sabha.
\item Facts:  P Nagarajan | position held | Member of the 16th Lok Sabha : P Nagarajan | occupation | politician.
\item For the given sentence, the information that the subject is a politician isn't written, but we can say that a Member of Parliament will be a politician, hence we mark it.
\item As another example, consider a sentence that says that a person did her Masters in Geography but doesn't explicitly mention her occupation directly. Still, we can mark the occupation=geographer fact as valid.
\end{itemize}
\end{itemize}
\begin{itemize}
\item If information in the fact requires external world knowledge, we do not mark that fact. 
\begin{itemize}
\item Sentence: Amruta's mother is a Malayali and her father is a Punjabi, and she was born in Mumbai.
\item Fact: Place of Birth | Chembur.
\item Even if you know that Chembur is in Mumbai, please don't mark it.
\end{itemize}
\end{itemize}
\item If some facts contain redundant information , then don’t mark it.
\item Abbreviations: If the part of the sentence is abbreviated in the facts or if the part of fact is abbreviated in the sentence, we don't consider those facts.
\begin{itemize}
\item Sentence: Field Marshal Archibald Percival Wavell, 1st Earl Wavell, GCB, GCSI, GCIE, CMG, MC, KStJ, PC (5 May 1883 – 24 May 1950) was a senior officer of the British Army and an Indian Viceroy.
\item Facts: Archibald Wavell, 1st Earl Wavell | award received | Virtuti Militari
\end{itemize}
\item Fact Generalisation 
\begin{itemize}
    \item If specific information is present in the sentence but there isn't an exact match in the fact list, then select the apt synonyms.
    \begin{itemize}
        \item Sentence: He had established the image of a good litterateur through his poems.
        \item Now if the fact list contains occupation as poet, and there is no other fact with occupation as litterateur, we consider the apt synonym and mark this fact as valid. 
    \end{itemize}
    \item If facts contain more specific terms as compared to the term present in the sentence then consider that fact for annotation (facts can contain more specific information).
    \begin{itemize}
        \item Sentence: Rajagopal Chidambaram (born 12 November 1936), commonly known as R. Chidambaram, is a Padma Vibhushan honored Indian scientist.
        \item Fact: Rajagopala Chidambaram | occupation | nuclear physicist
        \item We mark this fact as a nuclear physicist is also a scientist. The fact has more specific information and we mark it as valid.  
    \end{itemize}
\end{itemize}
\end{itemize}
\section{Case Studies}
Table~\ref{tab:caseStudies} shows examples of text generation from our mT5 model along with reference text. We choose one example per language from the test dataset of our \textsc{XAlign} dataset.

\begin{table*}[!t]
    \centering
    \scriptsize
    \begin{tabular}{|l|p{0.65in}|p{0.65in}|p{0.65in}|p{0.65in}|p{0.65in}|p{0.65in}|p{0.65in}|p{0.65in}|}
    \hline
    No.&hi&mr&te&ta&en&gu&bn&kn\\
    \hline
    \hline
    1&occupation&occupation&occupation&occupation&occupation&occupation&occupation&occupation\\
    \hline
    2&date of birth&date of birth&date of birth&position held&date of birth&date of birth&date of birth&cast member\\
    \hline
    3&position held&position held&position held&date of birth&position held&cast member&country of citizenship&date of birth\\
    \hline
    4&cast member&date of death&cast member&cast member&country of citizenship&position held&cast member&award received\\
    \hline
    5&country of citizenship&country of citizenship&date of death&country of citizenship&educated at&award received&member of sports team&position held\\
    \hline
    6&award received&place of birth&place of birth&educated at&date of death&date of death&date of death&date of death\\
    \hline
    7&place of birth&member of sports team&award received&place of birth&award received&languages spoken written or signed&educated at&performer\\
    \hline
    8&date of death&member of political party&member of political party&date of death&place of birth&place of birth&place of birth&place of birth\\
    \hline
    9&educated at&cast member&country of citizenship&award received&member of sports team&author&position held&author\\
    \hline
    10&languages spoken written or signed&award received&educated at&member of political party&member of political party&country of citizenship&award received&educated at\\
    \hline
    \end{tabular}
    \caption{Top-10 frequent fact properties across languages.}
    \label{tab:topkpredicates}
\end{table*}

\section{Details of F2T Alignment in \textsc{XAlign}}
\subsection{Stage 1: Candidate Generation}
Besides MuRIL, we also experimented with mBERT~\cite{devlin2019bert}, XLM-R~\cite{conneau2020unsupervised} and LaBSE~\cite{feng2020language}, but found MuRIL to perform the best on a small dataset of 500 examples, separately annotated for Stage-1 quality evaluation. The similarity score $sim(f_i,s_j)$ is  obtained as an average of the following 4 scores: MuRIL($f_i,s_j$), TFIDF-cos(translate($f_i$, $l$), $s_j$), TFIDF-cos($f_i$, translate($s_j$, English)), MuRIL(translate($f_i$, $l$), translate($s_j$, English)). For translating sentences, we use IndicTrans~\cite{ramesh2021samanantar}. When translating the facts, we retain the label of entities within the fact tuple for which Wikidata multi-lingual label is present in LR language, and we translate the remaining parts of the fact. 

We obtain $sim(f_i,s_j)$, i.e. a score between 0 and 1, for every (fact, sentence) pair. We filter out sentences if the most similar fact has similarity score less than a threshold $\tau$. By manual inspection, we fix $\tau=0.65$. For every remaining sentence, we retain at most top-$K$ facts sorted according to their scores. Empirically we observe that most sentences can be covered by less than 10 facts. Hence, we fix $K$=10. 

\subsection{Stage 2: Distant Supervision using KELM}
KELM is a distantly supervised dataset with automatically aligned (Wikipedia sentence, Wikidata facts) for English language. For a Wikipedia page corresponding to Wikidata entity $e$, a sentence $s$ is aligned with a Wikidata fact $f=\langle e, r, e'\rangle$ if $s$ contains subject $e$ and object $e'$.  

For every sentence in the dataset, we create a positive instance for every fact aligned with the sentence. For example, if sentence $s$ has two aligned facts $f_1$ and $f_2$, we create two positive instances. For every positive instance, we also create a negative instance as mentioned next. We order all the other sentences on the same Wikipedia page (which contains $s$) in decreasing order of semantic similarity and choose a sentence $s'$ randomly from top 10. We skipped top two sentences as they can be very similar to sentence $s$. We then use a fact extracted from sentence $s'$ along with the original sentence $s$ as a negative instance. We split the dataset in 90:10 for training and validation. Overall, the dataset contains 1,177,636 (54\% positive, 46\% negative) training instances and 130,849 (54\% positive, 46\% negative) validation instances.

\section{Further analysis of \textsc{XAlign} dataset}
\begin{figure}
    \centering
    \includegraphics[width=\columnwidth]{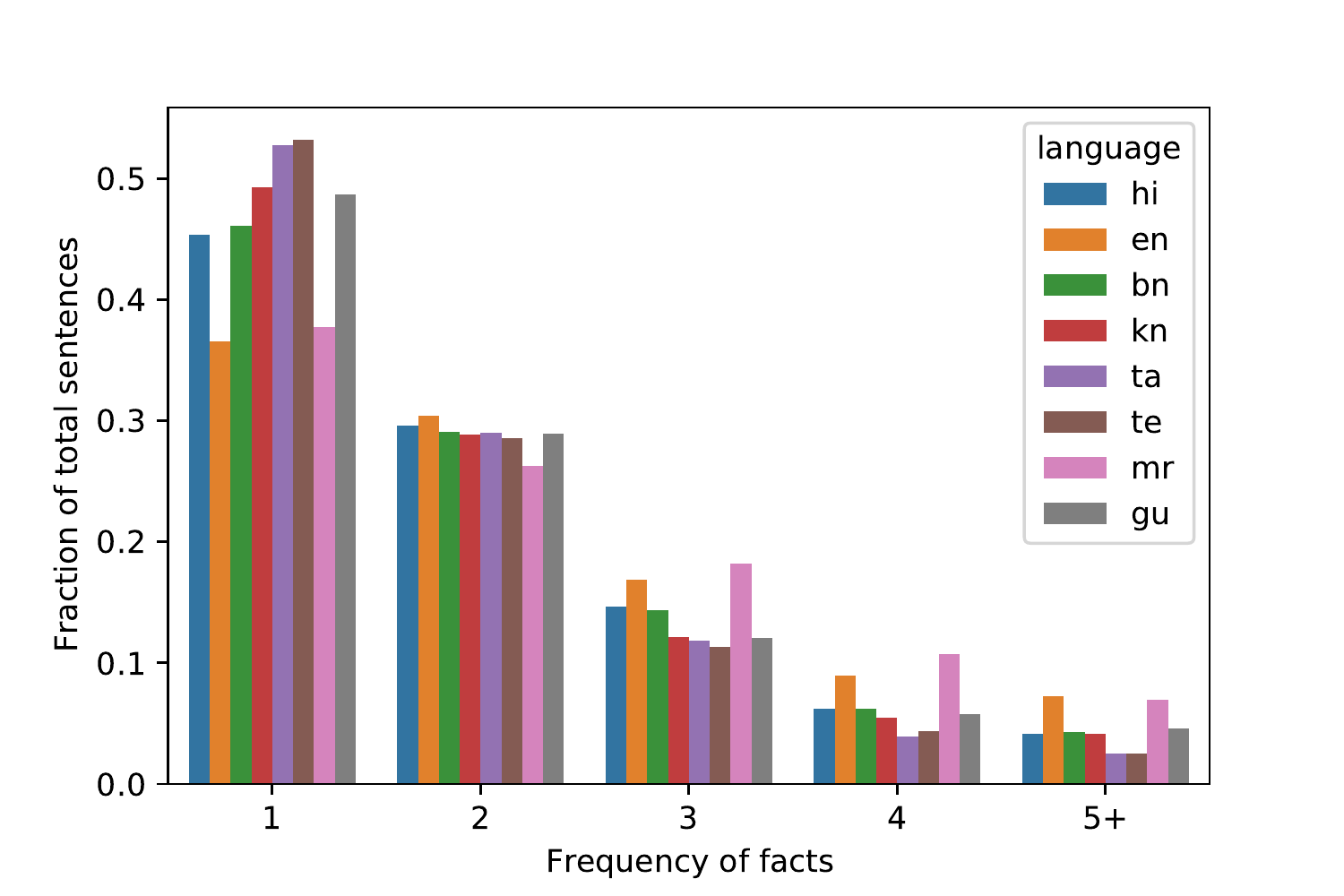}
    \caption{Fact Count Distribution across languages}
    \label{fig:language_fact_number_fraction}
\end{figure}

\begin{figure}
    \centering
    \includegraphics[width=\columnwidth]{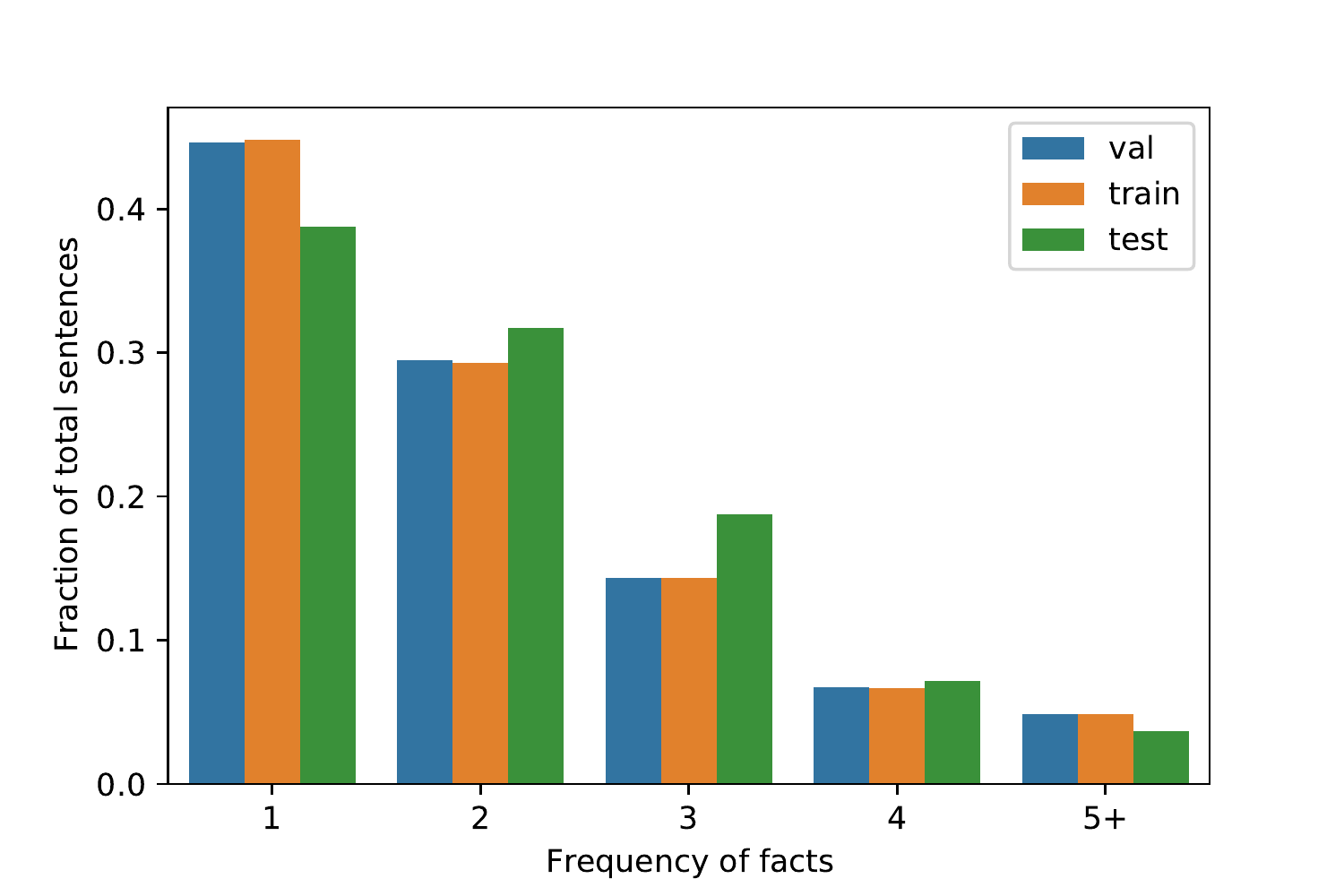}
    \caption{Fact Count Distribution across data subsets}
    \label{fig:train_test_val_factcount}
\end{figure}

Figs.~\ref{fig:language_fact_number_fraction} and~\ref{fig:train_test_val_factcount} show fact count distribution across languages and data subsets respectively. We observe that a large percent of sentences contain more than one fact across languages. Also, the distribution is similar across languages and data subsets.

Table~\ref{tab:topkpredicates} shows top 10 frequent fact properties across all the languages.

\section{Hyper-parameter Settings}
All experiments were run on a machine equipped with four
10GB RTX 2080 GPUs. For Stag-2: Distant supervision and Transfer learning approaches we have used large pretrained models publicly available on Huggingface~\cite{wolf2020transformers} library. For all experiments associated with it, we finetune for 5 epochs with L2-norm weight decay of 0.001 and dropout of 0.1. We set the learning rate of 1e-5, 2e-5 and 1e-3 for XLM-Roberta, MuRIL and mT5 respectively. We use batch size set of 32, 32 and 16 for XLM-Roberta, MuRIL and mT5 respectively. 

For all cross-lingual fact-to-text generation model excepts mT5 and translation baseline, we use a vocabulary
size of 64K subword learnt from training corpus using
SentencePiece~\cite{kudo2018sentencepiece} tokenizer. For transformer and GAT-Transformer model, we use 6 encoder and decoder layers, input embeddings of size 512 with 8 attention heads and
feedforward dimension of 2048. We optimized the
cross entropy loss using the AdamW optimizer. We use an initial learning rate of 1e-4, 4000 warmup steps and the learning rate annealing schedule as proposed in Vaswani et al.~\cite{vaswani2017attention}. We finetune the transformer and GAT-transformer models with batch size of 64 and 32 respectively for 100 epochs and early stopping with patience of 15. We finetune mT5-small model with constant learning rate of 1e-3, batch size of 2, weight decay 0.001 and dropout of 0.1. We optimize cross entropy loss using the Adafactor optimizer for 30 epochs.For all models, we use beam search with a beam size of 5 and length penalty set to 1. 

\section{Evaluation Metric}
For cross-lingual fact-to-text generation, we use overall BLEU scores for evaluating the multi-lingual models. We follow BLEU calculation steps mentioned in Ramesh et al.~\cite{ramesh2021samanantar} for English-Indic fact-sentence pairs. 
\end{document}